# Automated Object Behavioral Feature Extraction for Potential Risk Analysis based on Video Sensor

**Byeongjoon Noh[1] (powernoh@kaist.ac.kr), Dongho Ka[1] (kdh910121@kaist.ac.kr), Wonjun No[2] (jn0704@kaist.ac.kr) and Hwasoo Yeo[2]\* (hwasoo@kaist.ac.kr)**

[1]: Applied Science Research Institute, Korea Advanced Institute of Science and Technology, 291 Daehak-ro, Yuseung-gu, Daejeon, Republic of Korea
[2]: Department of Civil and Environmental Engineering, Korea Advanced Institute of Science and Technology, 291 Daehak-ro, Yuseung-gu, Daejeon, Republic of Korea

\**corresponding author*

*Abstract—* Pedestrians are exposed to risk of death or serious injuries on roads, especially unsignalized crosswalks, for a variety of reasons. To date, an extensive variety of studies have reported on vision-based traffic safety system. However, many studies required manual inspection of the volumes of traffic video to reliably obtain traffic-related objects' behavioral factors. In this paper, we propose an automated and simpler system for effectively extracting object behavioral features from video sensors deployed on the road. We conduct basic statistical analysis on these features, and show how they can be useful for monitoring the traffic behavior on the road. We confirm the feasibility of the proposed system by applying our prototype to two unsignalized crosswalks in Osan city, South Korea. To conclude, we compare behaviors of vehicles and pedestrians in those two areas by simple statistical analysis.  This study demonstrates the potential for a network of connected video sensors to provide actionable data for smart cities to improve pedestrian safety in dangerous road environments.

I. INTRODUCTION

Over the past decade, road traffic accidents globally posed a severe threat to human lives and have become a leading cause of premature deaths [1]. According to statistics, approximately 1.2 million people are killed each year from road traffic accidents, and up to 50 million are injured [2][3]. In particular, pedestrians are exposed to risk of death or serious injuries on the road for a variety of reasons, such as drivers failing to yield to pedestrians in crosswalks [2]. British Transport and Road Research Laboratory reported that crossing at an unsignalized crosswalk is as dangerous for pedestrians as crossing without a crosswalk or traffic signal [4]. International bodies such as World Health Organization (WHO) have recognized the seriousness of the problem. Therefore, it is necessary to alleviate deaths of vulnerable road users (VRUs).

In general, efforts that aim to advance vehicle safety systems can largely be divided into two types; 1) passive safety systems, which help reduce injuries when accidents occur, such as enhancement of vehicular safety system; and 2) active safety systems, which aim to prevent traffic accidents. These approaches forecast future driving states based on vehicle dynamics and traffic infrastructure.

Much research has been conducted on active safety systems. For example, the authors in [5] investigated the main factors contributing to the high fatality rates at pedestrian crosswalks in Poland. From trends in accident statistics from 2001 to 2013, they calculated level of safety at pedestrian crosswalks depending on road user behavior and adherence to traffic rules. As a result, they showed that factors such as speed limit, lighting conditions, and type of road contribute to the probability of pedestrian fatalities at unsignalized crosswalks.

The authors in [6] identified and compared the factors affecting pedestrian crash injury severity, such as average daily traffic, pedestrian age, and weather and vehicle type, at signalized and unsignalized roads. They used three years of pedestrian collision data from Florida, and focused on the relative severity of pedestrian injury rather than the absolute risk of vehicle-pedestrian crash. They recommend appropriate countermeasures to reduce the severity of pedestrian crashes such as improving lighting on urban corridors, placing standard crosswalks at unsignalized intersections, and blocking truck access at intersections with a high pedestrian concentration. Meanwhile, the authors in [3] diagnosed the infrastructure characteristics and deficiencies associated with pedestrian accidents at 95 urban locations. They concluded that the urban road network should be transformed to diminish the areas of car-pedestrian accidents and to significantly reduce vehicle speeds in areas vulnerable to traffic accidents.

These three studies focused on analyzing the vehicle dynamics and infrastructures in order to actively reduce traffic accidents. However, they only used historic accident data to determine how to transform the environment post facto. Alternatively, we can analyze factors for potential accident risk before traffic accidents occur, in order to prevent accidents proactively by deploying speed cameras, speed bumps, and other traffic calming measures. To date, an extensive variety of studies have reported on proactive traffic safety system, such as near-miss collision detection and risk analysis. In order to obtain the data required, such as vehicle and pedestrian positions and speed, vision-based traffic surveillance systems have been widely applied. Most studies focus on traffic-related objects' behavior analysis [2][7][8][9][10].

The authors in [7][8] proposed a novel framework to evaluate pedestrian safety at unsignalized roads, and investigated it during pedestrian-vehicle interactions based on vision-based trajectory data. In addition, the authors in [9] analyzed field traffic at unsignalized crosswalks in Beijing, China and Munich, Germany by using video recording and image processing, in order to understand intercultural differences in urban traffic behavior. They studied various pedestrian behavior factors such as walking phases, speed, and gap acceptance.

Moreover, these analyses can provide powerful and useful information for urban planners in improving the road environment. For example, researchers in [10] assessed collision risk between vehicles and pedestrians, using a variety of features such as pedestrian counts and automobile traffic flow. Across 488 intersections in Minneapolis, they identified a "safety in numbers" effect that reduced accident risk in spaces with high pedestrian activity, supporting efforts to improve walkability as a way to improve safety.

In vision-based traffic safety and surveillance systems, one important step is to obtain the object's behavioral factors from video. However, many studies required manual inspection of huge volumes of traffic video to reliably extract traffic-related object trajectories [11][12][13]. Since most vision sensors deployed on roads, such as closed circuit television (CCTV), have an oblique view of the vehicles, it is difficult to extract an object's precise coordinates and features without multiple sensors or complex data handling processes [14][15]. The authors in [14] used convolutional neural networks (CNNs), a popular deep learning model, to automatically construct 3-D bounding boxes around cars from only a single camera viewpoint. This allowed them to accurately place the car in space and identify its trajectory. Likewise, the authors in [15] demonstrated automatic pedestrian tracking over 2 hours of video data collected at a major signalized intersection in New York City. They used an automated video processing system provided by the University of British Columbia, which functioned to calibrate the image into the overhead view, and tracked the pedestrians using computer vision techniques.

Unlike the perspectives in the current research literature, in this study, we propose an automated and simpler system for effectively extracting object behavioral features from video sensors (*e.g.* CCTV) deployed on the road. The proposed system has three objectives; 1) To detect and segment traffic-related objects (vehicle and pedestrian) into individual objects; 2) To recognize the contact point of each object from an overhead perspective; and 3) To automatically extract their behavioral features such as vehicle speed and acceleration, or pedestrian speed. We conduct basic statistical analysis on these features such as correlation analysis, to show how they can be useful for monitoring the traffic behavior on the road. We confirm the feasibility of the proposed system by applying our prototype to two unsignalized crosswalks in Osan city, South Korea. To conclude, we compare behaviors of vehicles and pedestrians in those two areas by simple statistical analysis. This research is the first step in our long-term effort to develop an IoT system of smart cameras that help identify unsafe pedestrian environments in cities.

## II. Materials and methods

### A. Data Source

In our experiments, we used video data from CCTV cameras deployed on two roads in Osan city, Republic of Korea; 1) Segyo complex #9 back gate #2 (spot A); and 2) Noriter daycare #2 (spot B). These CCTVs are deployed over unsignalized crosswalks, and are intended to record and deter instances of street crime. All video frames were processed locally on a server we deployed in the Osan smart city integrated operations center, and we only retained and viewed the processed trajectory data, after deleting the original video frames. Future systems would employ internet-connected cameras that process images on-device in real-time, and transmit only trajectory information back to servers.

Figure 1 shows the camera views which were at oblique angles above the road. Figure 2 illustrates the roads, sidewalks, and crosswalks at spots A and B from overhead perspective, along with a sample of object trajectories that resulted from our processing. Blue lines are trajectories of pedestrians, and green lines are vehicles. The widths of both crosswalks are about 15m, and speed limits on surrounding roads are 30km/h. In addition, spot A is nearby a high school but is not designated a school zone, whereas spot B is a school zone. In Korea, school zones are certain roads near facilities for children under age 13, including elementary schools, daycare centers, and tutoring academies. In such areas, road safety features are deployed to ensure safe movement for children to prevent traffic accidents. Moreover, drivers who have accidents or break the rules in these areas receive heavy penalties, such as doubling of fines. As seen in Figure 1 (b), the road is covered with red urethane, and a fence separates the road from the sidewalk.

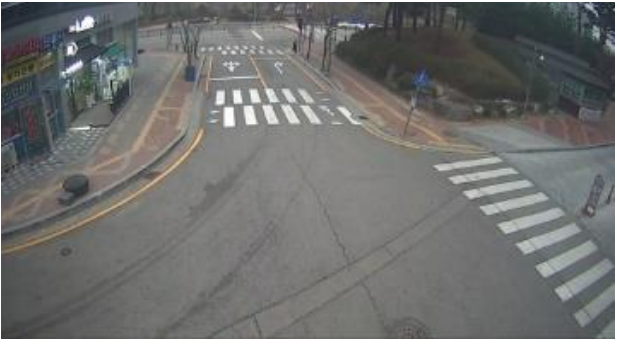 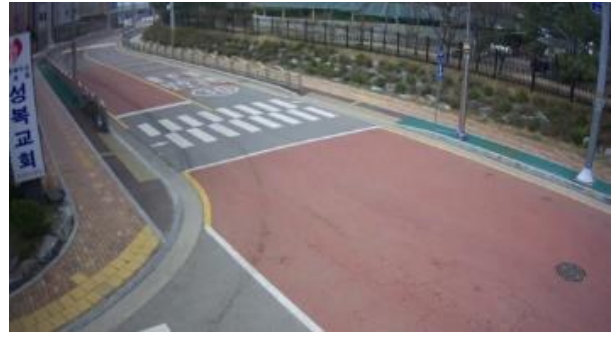

(a)                  (b)

**Fig. 1.** Actual CCTV views in (a) spot A and (b) spot B

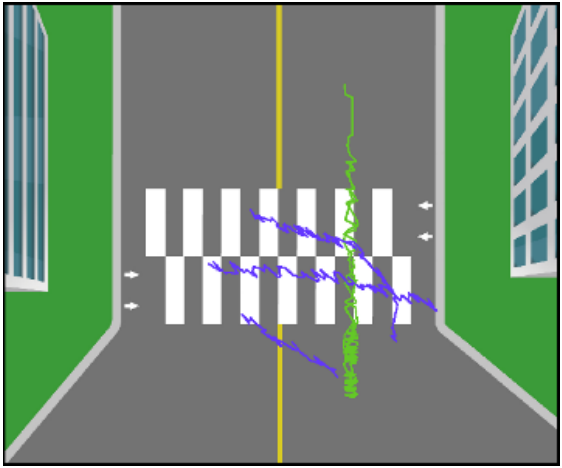 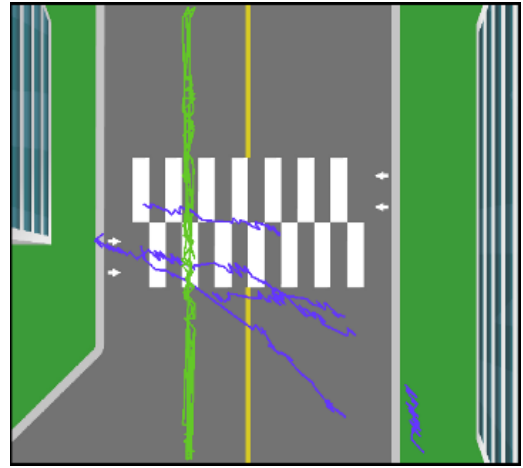

(a)                  (b)

**Fig. 2.** Overhead view diagram with sampled object trajectories

Frame sizes from recorded video, at both spots, are 1,280 x 720 pixels. Videos had been recorded at 15 FPS (frames-per-sec) and 11 FPS in spots A and B, respectively. Since these areas are located near schools and residential complexes, the "floating population" passing through there is high, especially during rush hour. Thus, we used video recorded between 8am and 9am on weekdays. Next, we extracted only video clips containing scenes where at least one pedestrian and one car were in camera view at the same time. As a result, we processed 429 and 359 video clips of potential pedestrian-car interaction events in spots A and B, respectively.

*B. Object Detection and Overhead Contact Point Recognition*

In this section, we describe how we detected the traffic-related objects such as vehicle and pedestrian. Since video data was obtained in oblique view, the "contact points" of each object (reference points for determining their speed and distance from each other) depended on their position and direction, and angle of the camera. Thus, it is important to obtain the contact point of each object and convert to overhead perspective to extract its precise behavioral features.

First, we resampled each video, using every 5th frame. Then, we used a pre-trained mask R-CNN model to detect and segment objects in the frame. This model is an extension of faster R-CNN, and provides the output in the form of bitmap mask as well as bounding-box [16]. Until now, deep learning algorithms have dominated various computer vision tasks such as object detection and instance segmentation [17]. In particular, faster R-CNN has been widely applied to recognize and classify the objects in image frame [18]. However, since faster R-CNN normally outputs only a bounding-box, we used mask R-CNN to generate a segmentation mask over the Region of Interest (RoI), which could be combined with the bounding box to estimate the contact point of vehicles and pedestrians [19].

In our experiment, we used the object detection API (Application Programming Interface) implemented in Tensorflow platform and the pre-trained mask R-CNN model (ResNet-101-FPN) by Microsoft common objects in context (MS COCO) image dataset [20][21]. Note that the goal of this module was to detect only vehicles and pedestrians, which it did with approximately 99.9% accuracy. Thus, we did not need to train the model further for our purposes.

In order to recognize the contact points of vehicles and pedestrians, we first needed to capture "ground tip" points of these objects. Ground tip means a point on the ground directly underneath the front center of the object, in the oblique view. For example, ground tip of a car is located directly under the center of its front bumper, and ground tip of pedestrian is located on the ground between the feet.

For vehicles, we used the object mask and central axis line of the vehicle lane to determine the ground tip; more detailed procedure for this is described in our previous study, [22]. We used a similar procedure for pedestrians. From their tiptoe points in mask, as seen in Figure 3, we determined the midpoint close to the ground.

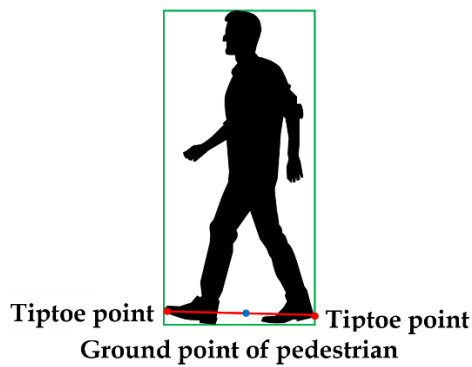

**Fig. 3.** Ground point of pedestrian for recognizing its overhead contact point

Then, we transformed these points from the oblique camera perspective to an overhead perspective using the "transformation matrix" function in OpenCV library. Transformation matrix can be derived from four pairs of corresponding "anchor points" in real image and virtual image. In our experiment, we measured the real length and width of the rectangular crosswalk area, then redrew it in virtual space from an overhead perspective with proportional dimensions and oriented orthogonally to the x-y plane. Then, we identified the vertices of the rectangular crosswalk area in the camera image's oblique view and the corresponding vertices in our redrawn overhead view, and used these to initialize the OpenCV function to apply to all detected contact points.

With these contact points, we can extract the object's precise behavioral features such as speed and acceleration.

*C. Object Behavioral Feature Extraction*

In this section, we describe how to extract the object's behavioral features from the recognized overhead contact point. In order to extract a single object's behavioral features, it is important to distinguish each object captured in consecutive and multiple frames. Thus, we apply a simple tracking algorithm in our experiment by using threshold and minimum distance methods. Much research has been conducted on object tracking using vision sensor, with crowded scenes requiring more complex tracking algorithms [23][24][25]. In our experiment, since most unsignalized crosswalks are on narrow roads with light pedestrian traffic, we applied a simpler, low-computation tracking algorithm [23]. The core of this method is to use distance threshold and minimum distance of objects between consecutive frames.

Each frame includes multiple object positions as x and y coordinates (contact points), and each position has a unique identifier ordered by detection accuracy. For example, Figure 4 (a) represents detected object (pedestrian) positions in two consecutive frames. The first frame has three positions named *A*, *B* and *C*, and second frame has two positions named *D* and *E*. Assume that Figure 4 (b) shows the actual movement of pedestrians between the two frames. Two pedestrians start at positions *B* and *C* in the first frame and move to positions *D* and *E* in the second frame, respectively. The third pedestrian moves from *A* to somewhere out-of-frame.

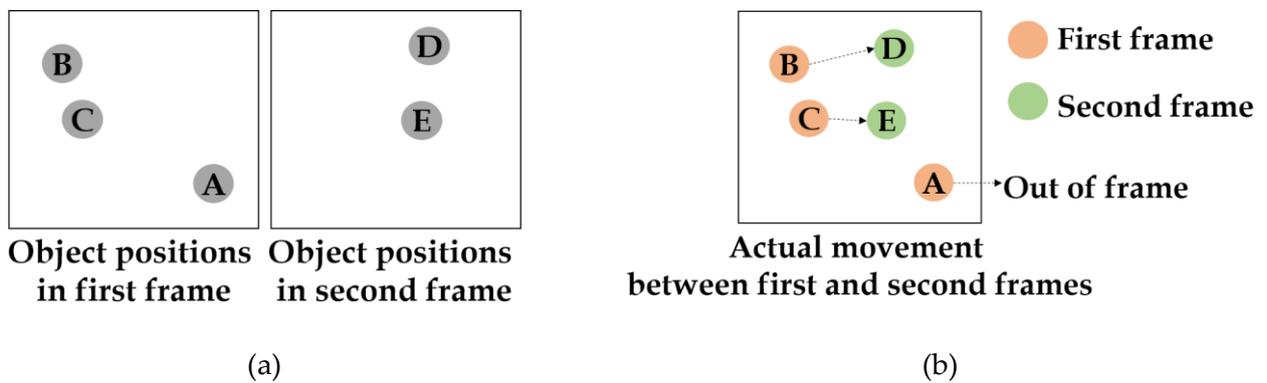

**Fig. 4.** Example of object positions in two consecutive frames

To infer the trajectories of each object, we set frame-to-frame distance thresholds for vehicles and pedestrians, and compare all distances between positions from first to second frame as seen in Figure 5. In this example, if we set the pedestrian threshold at 2.0, *A* is too far from either position in the second frame, so we assume it has moved out-of-frame. When *B* is compared with *D* and *E*, it is closer to *D*; Likewise, *C* is closer to *E*. We infer that a pedestrian at *B* moved to *D*, and one moved from *C* to *E*, while a pedestrian at *A* left the frame. We apply this algorithm to each pair of consecutive frames in dataset to rebuild the full trajectory of each object.

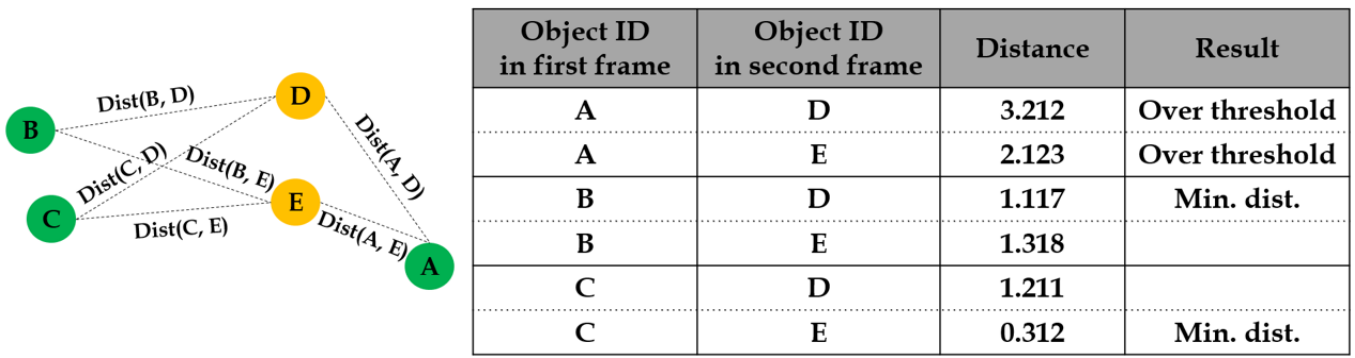

| Object ID in first frame | Object ID in second frame | Distance | Result |
|---|---|---|---|
| A | D | 3.212 | Over threshold |
| A | E | 2.123 | Over threshold |
| B | D | 1.117 | Min. dist. |
| B | E | 1.318 | |
| C | D | 1.211 | |
| C | E | 0.312 | Min. dist. |

**Fig. 5.** Example of trajectory algorithm

Afterwards, we extract four behavioral features for each object: vehicle speed, pedestrian speed, vehicle acceleration, and distance between vehicle and pedestrian.

Vehicle and pedestrian speed: Vehicle and pedestrian velocities are basic measurements that can signal potential risky situations. Speed limit in our testbed is 30*km/hour*; if there are many vehicles detected traveling over the limit at any point, that contributes to the risk at that location. Pedestrian speed alone is not an obvious signal for risk, but when analyzed together with other features, we may find important correlations and interactions between them.

To extract their velocities, we simply divide an object's distance traveled between frames by the time interval between frames. Videos in spots A and spot B were recorded at 15 FPS and 11 FPS, respectively, and we sampled every 5th frame. Thus, in spot A, the time interval *F* between two consecutive frames was 1/3 sec, and in spot B, 5/11 sec.

Meanwhile, we had to convert from pixel distance in our transformed overhead coordinates to real-world distance in meters. We derived this pixel-per-meter constant (*P*) using the length of the crosswalks as our reference point. In our experiment, we measured the actual lengths of both crosswalks in spots A and B as 15m, and pixel length of crosswalk as 960 pixels. Thus, 1 meter is about 46 pixels (= 960 / 15) in both spots. Finally, object's speed can be calculated as following:

$$Velocity = \frac{pixel\ distance}{F * P}\ (m/s)$$

The unit is finally converted into $km/h$.

Vehicle acceleration: Vehicle acceleration is also an important factor to determine potential risk. If many vehicles maintain speed or accelerate while approaching the crosswalk, it can be regarded as a dangerous situation. Ideally, we would see some deceleration (negative values) near the crosswalk.

To extract vehicle's acceleration, we use the difference between vehicle's velocities in current frame ($v_0$) and in next frame ($v$):

$$Acceleration = \frac{v - v_0}{F}\ (m/s^2)$$

The unit is finally converted into $km/h^2$.

Distance between vehicle and pedestrian: This feature means distance between vehicle and closest pedestrian. In general, if the distance between vehicle and pedestrian crossing the crosswalk is short, the driver should slow down with more caution. Otherwise, this presents a potentially risky situation.

$$Distance = \frac{object\ distance}{P}\ (m)$$

## III. EXPERIMENTS AND RESULTS

With the proposed system, each object's behavioral feature can be extracted automatically. We applied this system to two unsignalized crosswalks in Osan city, Korea. Then, we conducted statistical analysis on the extracted features, and compared behavioral characteristics of objects between the two regions.

## A. Experimental Design

In this section, we briefly explain the result of preprocessing and experimental design for analysis. Total numbers of records are 2,635 and 1,400 frames from videos in spots A and spot B, respectively. Through preprocessing, we eliminated outlier frames based on extreme feature values, yielding 2,304 and 991 frames, respectively. Table I describes the preprocessed dataset with maximum, minimum, and average values of each feature. As seen in Table I, the maximum speed of vehicles was 89.6km/h and 88.3km/h in spots A and B, respectively. The speed limit at these two spots is 30km/h, so both spots had at least one case of dangerous speeding during the study period.

TABLE I. FEATURES INFORMATION IN SPOTS A AND B

| Object | Statistical Measurement | Spot A | Spot B |
|---|---|---|---|
| Vehicle | Min. speed ($km/h$) | 0.0 | 0.0 |
| | Max. speed ($km/h$) | 89.6 | 88.3 |
| | Avg. speed ($km/h$) | 25.0 | 26.1 |
| | Min. acceleration ($km/h^2$) | -11.9 | -9.9 |
| | Max. acceleration ($km/h^2$) | 9.9 | 10.0 |
| | Avg. acceleration ($km/h^2$) | -0.28 | -0.1 |
| Pedestrian | Min. speed ($km/h$) | 0.0 | 0.0 |
| | Max. speed ($km/h$) | 14.4 | 14.7 |
| | Avg. speed ($km/h$) | 2.4 | 2.6 |

In our experiments, we compared the behavioral features of two spots by using simple statistical methodologies such as histogram and boxplot analysis. Next, by conducting correlation analysis, we can obtain the relationships between each feature. For this, we performed [0, 1] normalization on the features as follows:

$$d_i = \frac{d_i - \min(d)}{\max(d) - \min(d)}$$

## B. Results and discussions

In this section, we describe the results of histograms, boxplots, and correlation analysis. Figure 6 (a) ~ (c) and Figure 7 (a) ~ (c) illustrate histograms and boxplots of each feature in two posts, respectively. As seen in Figure 6 (a) and 7 (a), the overall distributions of vehicle velocities are similar with each other in spots A and B, and skewed low since many cars stop or drive slowly in these areas.

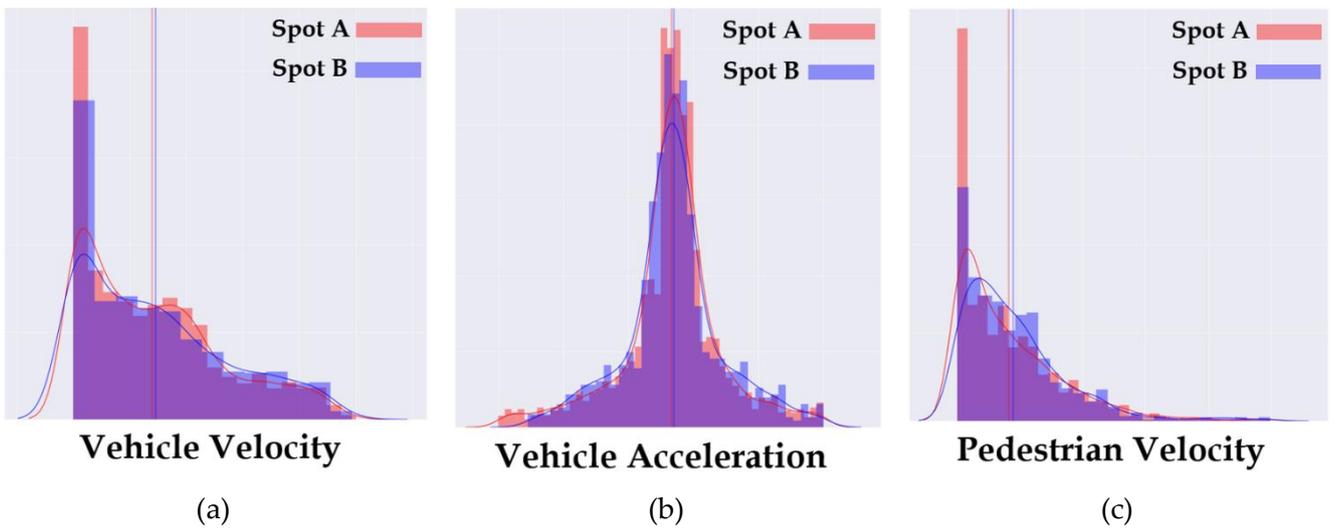

**Fig. 6.** Results of histograms in spots A and B

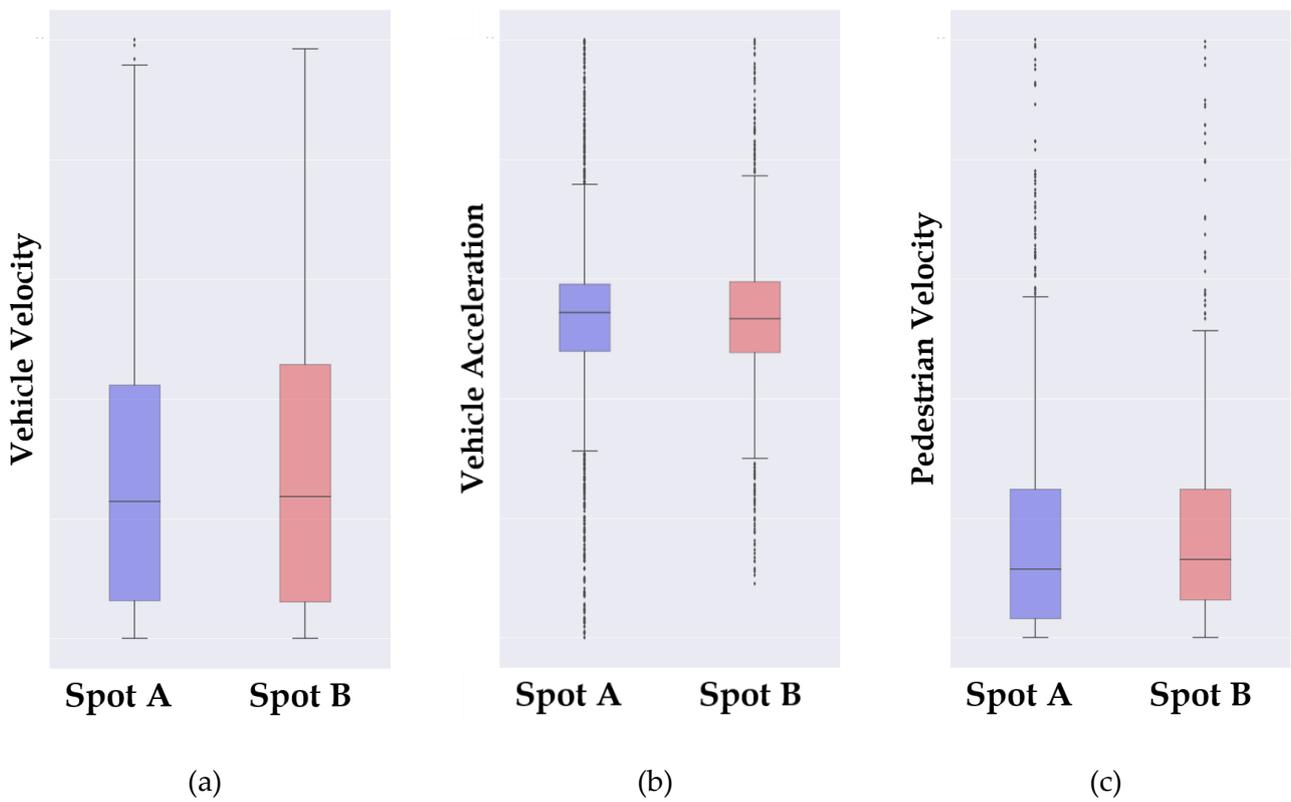

**Fig. 7.** Results of boxplot analysis in spots A and B

Figure 6 (b) and 7 (b) show the distribution and boxplot of vehicle accelerations. Vehicle accelerations mostly stayed close to zero. However, in spot A, there is a longer tail on the deceleration side, which might mean a few cars are suddenly slowing down to avoid pedestrians. Meanwhile, in both areas, some vehicles showed high acceleration. This could be interpreted in two ways: either some cars are accelerating to higher speeds through the crosswalk, or are accelerating more quickly after coming to a stop prior to the crosswalks. The reason for the difference in spots A and B needs to be analyzed from the perspective of urban planning, and addressed by decision makers.

As shown in Figure 6 (c) and 7 (c), average pedestrian speed is also similar with each other in spots A and B, and skewed low since most pedestrians walk slowly in crosswalk, or stop to wait for cars to pass by. However, some pedestrians run with high speed. It may be related to the vehicle speed, if, for example, pedestrians are running to avoid fast-moving cars. Pedestrian speed alone cannot determine the reason for these results.

By conducting correlation analysis, we can study the relationships between each feature. Figure 8 and 9 represent correlation matrices in spots A and B, respectively. In spot A, there is negative correlation between vehicle speed and vehicle-pedestrian distance, implying that vehicles are moving faster when closer to pedestrians; spot B appears to be safer in that respect, with vehicle speeds slower when closer.

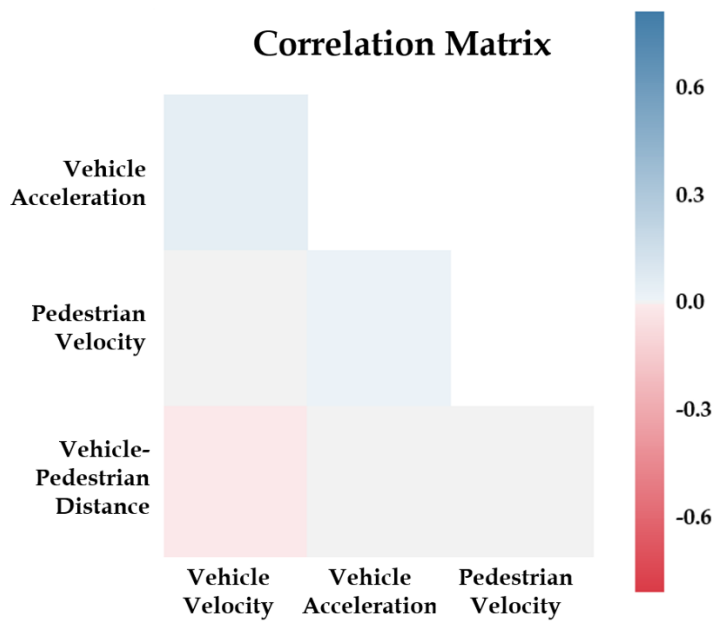

**Fig. 8.** Result of correlation analysis in spot A

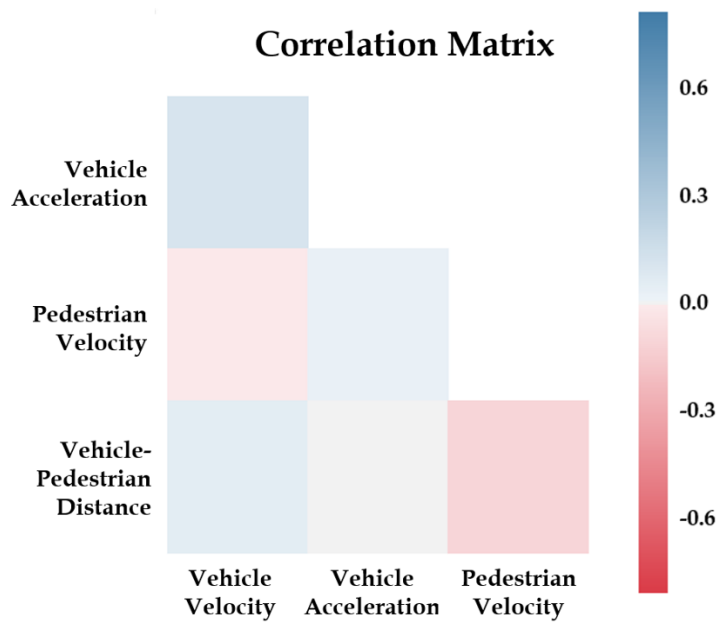

**Fig. 9.** Result of correlation analysis in spot B

In spot B, we can observe negative correlation between pedestrian speed and vehicle-pedestrian distance. This case could result from two behaviors: pedestrians running fast when in the path of an approaching car, or pedestrians slowing or stopping to wait well before the cars reach the crosswalk. Unlike spot A, spot B shows negative correlation between vehicle speed and pedestrian speed. In other words, when the car is moving fast, pedestrian is slow or stopped, and vice versa. Remarkably, in both areas, there is positive correlation between vehicle speed and acceleration, especially in spot B. It means that despite the presence of pedestrians, vehicles already traveling at high speeds were more likely to accelerate.

IV. Conclusion

In this paper, we proposed an automated and simpler system for effectively extracting traffic object behavioral features based on video. The core methodologies are 1) to recognize the objects' precise contact points by modified deep learning object detection algorithm; and 2) to extract their behavioral features by frame. We validated the feasibility and applicability of the proposed system by developing the prototype using Tensorflow platform and OpenCV library, and applying it to an actual road operating CCTV in Osan city, South Korea. We believe that the proposed system can be a useful supplementary tool and help decision makers to identify street locations where events endangering pedestrians frequently occur. Scaled up, the system could use many distributed, internet-connected cameras to collect such behavior information throughout a smart city, while protecting privacy by retaining only the trajectories of vehicles and pedestrians, without any identifying information.

The proposed system itself does not identify the best control or traffic calming measures to prevent traffic accidents. However, it provides clues to support further investigation using large datasets of vehicle and pedestrian behavior processed automatically. Thus, traffic engineers and urban designers must collaborate using these clues to improve the safety of the space. Our next steps are to develop methodologies that provide greater context and insight for such collaborations, so that we can guide urban design decisions using objective data analysis.